\documentclass[11pt]{article}

\usepackage[final]{acl}

\usepackage{times}
\usepackage{latexsym}

\usepackage[T1]{fontenc}

\usepackage[utf8]{inputenc}
\usepackage{amssymb}
\usepackage{microtype}
\usepackage{listings}
\usepackage{xcolor}
\usepackage{booktabs}
\usepackage{inconsolata}

\usepackage{graphicx}
\usepackage{bbm}
\usepackage{multirow}
\usepackage{array}
\usepackage{xspace}
\usepackage{bbding}
\usepackage{makecell}
\usepackage{enumitem}
\usepackage{pifont}
\usepackage{tcolorbox}

\newcommand{\minosraw}{Minos-Raw\xspace}
\newcommand{\minosfinal}{Minos-57K\xspace}

\title{\textsc{Minos}: A Multimodal Evaluation Model for Bidirectional Generation Between Image and Text}

\author{Junzhe Zhang\textsuperscript{1}, Huixuan Zhang\textsuperscript{1}, Xinyu Hu\textsuperscript{1}, Li Lin\textsuperscript{1}\\ \textbf{Mingqi Gao\textsuperscript{1}, Shi Qiu\textsuperscript{2}, Xiaojun Wan\textsuperscript{1}}\\ 
\textsuperscript{1}Wangxuan Institute of Computer Technology, Peking University \\
\textsuperscript{2}School of Physics, Peking University\\
         \{junzhezhang, zhanghuixuan, efsotr\_l, shiqiu\}@stu.pku.edu.cn  \\ \{huxinyu, gaomingqi, wanxiaojun\}@pku.edu.cn}

\begin{document}
\maketitle
\begin{abstract}
Evaluation is important for multimodal generation tasks, while traditional multimodal evaluation metrics suffer from several limitations.
With the rapid progress of MLLMs, there is growing interest in applying MLLMs to build general evaluation systems. 
However, existing researches often simply collect large-scale evaluation data for training, while overlooking the quality of evaluation data. What's more, current proposed evaluation models often struggle to achieve consistently strong performance across both image-to-text (I2T) and text-to-image (T2I) tasks.
In this paper, through rigorous quality control strategies, we construct a comprehensive multimodal evaluation dataset, \minosfinal, with evaluation samples across 15 datasets, for developing the multimodal evaluation model Minos with SFT and preference alignment training strategies. 
Notably, despite using less than half the scale of the training data of prior work, our model achieves state-of-the-art evaluation performance across 16 out-of-domain datasets covering both I2T and T2I tasks among all open-source multimodal evaluation models and remain competitive with closed-source models. Extensive experiments demonstrate the importance of leveraging quality control process, jointly training on evaluation data from both I2T and T2I generation tasks and further preference alignment.
\end{abstract}

\section{Introduction}

Multimodal evaluation\cite{huang2024surveyevaluationmultimodallarge, zhang2023gpt4visiongeneralistevaluatorvisionlanguage, ge2023mllm} is crucial for multimodal generation tasks and developing multimodal models. A reliable evaluation not only enables more accurate comparison across models, but also plays a crucial role in the development of multimodal models.
Although multimodal evaluation is crucial, traditional metrics still face notable limitations\cite{hessel2022clipscorereferencefreeevaluationmetric, manas2024improving}, such as correlating poorly with human judgments,  requiring reference data, and task-dependent. Therefore, developing a general-purpose multimodal evaluation system is becoming increasingly important.

With the rapid development of multimodal large language models (MLLMs), recent studies\cite{chen2024mllm, lee2024prometheus, xiong2024llavacritic, wang2025unifiedrewardmodelmultimodal} begin to explore applying MLLMs for building general multimodal evaluation systems. For example, LLaVA-Critic\cite{xiong2024llavacritic} collects existing multimodal generation data and prompts the GPT-4o\cite{openai2024gpt4technicalreport} to obtain evaluation results among various image-to-text(I2T) generation tasks to train a general evaluation model. UnifiedReward\cite{wang2025unifiedrewardmodelmultimodal} applies data of LLaVA-Critic and other open-source multimodal evaluation data to train a unified multimodal reward model across multiple modalities. 
However, they simply collected large-scale evaluation data from diverse tasks for training, without subjecting it to rigorous quality control.

We argue that multimodal evaluation models should be developed using high-quality evaluation data curated through a strict and systematic quality control process, which is as important as the task diversity of evaluation data.
Following this insight, we first try to collect existing human-annotated evaluation datasets on multi-directional multimodal generation tasks including both I2T and T2I tasks as high-quality evaluation data sources. 
However, human-annotated evaluation data remain scarce and often provide only evaluation scores without clear analyses, and many multimodal tasks even lack human evaluation data. Therefore, we manually define evaluation guidelines and annotate additional evaluation data for more multimodal tasks. 
After obtaining the annotated large-scale evaluation data, different from previous researches\cite{xiong2024llavacritic, wang2025unifiedrewardmodelmultimodal}, instead of directly using them for training, we conduct strict quality control from both the instance-level and dataset-level perspectives to construct our evaluation dataset \minosfinal. 
\minosfinal is a multimodal evaluation dataset constructed from 15 different multimodal datasets including both I2T and T2I. 



\begin{table*}[htp]
    \centering
    \setlength{\tabcolsep}{2mm}{
    \renewcommand\arraystretch{1.1}
    \begin{tabular}{cccccc}
    \toprule 
    \multirow{2}*{Method} & Task              & Task-Specific           & Instance-level    & Dataset-level        & Preference    \\
                         & Diversity          & Guideline               & Filtering         & Balancing            & Alignment     \\\hline 
    Prometheus-Vision    & I2T                & Generated               & \ding{55}         & \ding{51}            & \ding{55}     \\ 
    LLaVA-Critic         & I2T                & \ding{55}               & \ding{55}         & \ding{55}            & \ding{55}     \\
    UnifiedReward        & Multi-Directional  & No I2T                  & \ding{55}         & \ding{55}            & \ding{55}     \\
    Minos                & Multi-Directional  & \ding{51}               & \ding{51}         & \ding{51}            & \ding{51}     \\
    \bottomrule
    \end{tabular}}
    \caption{Comparison of Minos and previous researches of mainstream multimodal evaluation models.}
\label{table:comparison}
\end{table*}

Moreover, prior trained multimodal evaluation models such as LLaVA-Critic and UnifiedReward typically rely solely on supervised fine-tuning(SFT) using evaluation data, while overlooking the alignment stage\cite{ouyang2022traininglanguagemodelsfollow, lee2023rlaif} that plays a crucial role in MLLMs development. 
Inspired by this, we further construct multimodal preference dataset, thereby enabling Direct Preference Optimization(DPO)\cite{rafailov2023direct} alignment of the multimodal evaluation model. 
Building upon the SFT evaluation data quality control strategy, we further apply more strict filtering to obtain the high-quality evaluation preference dataset Minos-DPO-5.8K.

After SFT on \minosfinal and DPO alignment on Minos-DPO-5.8K with Qwen3-VL-8B\cite{bai2025qwen3vltechnicalreport} backbone, we obtain our multimodal evaluation model Minos.
Our proposed Minos is capable of evaluating diverse multimodal generation tasks across T2I and I2T tasks~(\textbf{Modality Generalization}) by providing reference-free scores~(\textbf{Independence}) and generating human-interpretable analysis~(\textbf{Interpretability}), highlighting its practical value. Minos outperforms previous open-source state-of-the-art evaluation models, and is competitive or even surpasses the performance of some closed-source models. Our experiments further demonstrate that, evaluation data quality and task diversity are the key factors influencing multimodal evaluation capability.

Overall, our main contributions are as follows\footnote{Our code is available at \url{https://github.com/reroze/MINOS}}:
\begin{itemize}
    \item We construct \textbf{\minosfinal}, a large-scale multimodal evaluation dataset across 15 datasets, including both T2I and I2T generation tasks, with rigorous construction procedure and strict filtering process. We further construct high-quality multimodal evaluation preference dataset \textbf{Minos-DPO-5.8K}. 
    \item We propose Minos, a multimodal evaluation model with modality generalization, independence and interpretability, trained using SFT and DPO alignment strategy on \minosfinal and Minos-DPO-5.8K. Minos achieves \textbf{state-of-the-art} performance among all open-source evaluation models, and is competitive or even surpasses the performance of closed-source models.
    \item Extensive experimental results demonstrate that the quality, task diversity of evaluation data, preference alignment, and evaluation analysis together contribute to improved evaluation capability in evaluation models.
\end{itemize}

\section{Related Work}
\subsection{MLLM as a Judge}
As multimodal large models (MLLMs) are increasingly employed to construct evaluation metrics across various tasks\cite{huang2024surveyevaluationmultimodallarge, xia2024mmie}, building a unified evaluation model for multiple multimodal tasks based on MLLMs has become a promising direction. The MLLM-as-a-Judge\cite{chen2024mllm} benchmark provides human-annotated evaluation data spanning 14 tasks and evaluates the performance of both open-source and closed-source MLLMs as evaluation models. Prometheus-V\cite{lee2024prometheus} was the first to leverage MLLM to construct a dedicated multimodal evaluation model. LLaVA-Critic\cite{xiong2024llavacritic} further collected a range of pairwise and pointwise evaluation data across multiple tasks annotated by GPT, and trained a larger-scale multimodal evaluation model. 
Recent studies have proposed several multimodal reward models\cite{zhang2025r1rewardtrainingmultimodalreward, wang2025unifiedrewardmodelmultimodal}, which are predominantly designed for pairwise preference evaluation. Among them, UnifiedReward natively supports both pointwise and pairwise evaluation. 
However, for the developoment of these pointwise evaluation models, they often simply applying evaluation data for training without strict filtering process. Existings multimodal evaluation models\cite{lee2024prometheus, xiong2024llavacritic, wang2025unifiedrewardmodelmultimodal} lack further DPO alignment after SFT stage, which could further improve the evaluation capacity of model. Main difference between Minos and previous evaluation models are listed in table \ref{table:comparison}.

\subsection{Multimodal Human Evaluation}

With the rapid advancement of multimodal research, increasing attention has been paid to the evaluation of multimodal tasks. Early works, such as CLIP-Score\cite{hesselclipscore}, introduced human evaluation datasets for image captioning. However, these early datasets often exhibited diverse and inconsistent formats, making them difficult to consolidate into a unified training resource for evaluation models. More recently, several studies\cite{wada2024polos, manas2024improving, xu2023imagereward, liang2024richhumanfeedbacktexttoimage} have collected human-annotated multimodal evaluation data across a variety of image-to-text and text-to-image tasks. For I2T evaluation, Polaris\cite{wada2024polos} introduced the image captioning dataset, comprising 131k human ratings annotated by 550 unique annotators. Similarly, LAVE\cite{manas2024improving} proposed a human evaluation dataset for visual question answering, which includes 29k human-labeled instances. On the T2I side, datasets such as ImageReward\cite{xu2023imagereward}, RichHF-18K\cite{liang2024richhumanfeedbacktexttoimage} and GenAI-Bench\cite{jiang2024genai} have been developed to support human evaluation. Nonetheless, despite these efforts, high-quality and large-scale human evaluation datasets remain limited, many multimodal tasks still lack sufficient human evaluation data.

\begin{figure*}[htp]
    \centering
    \includegraphics[width=15cm]{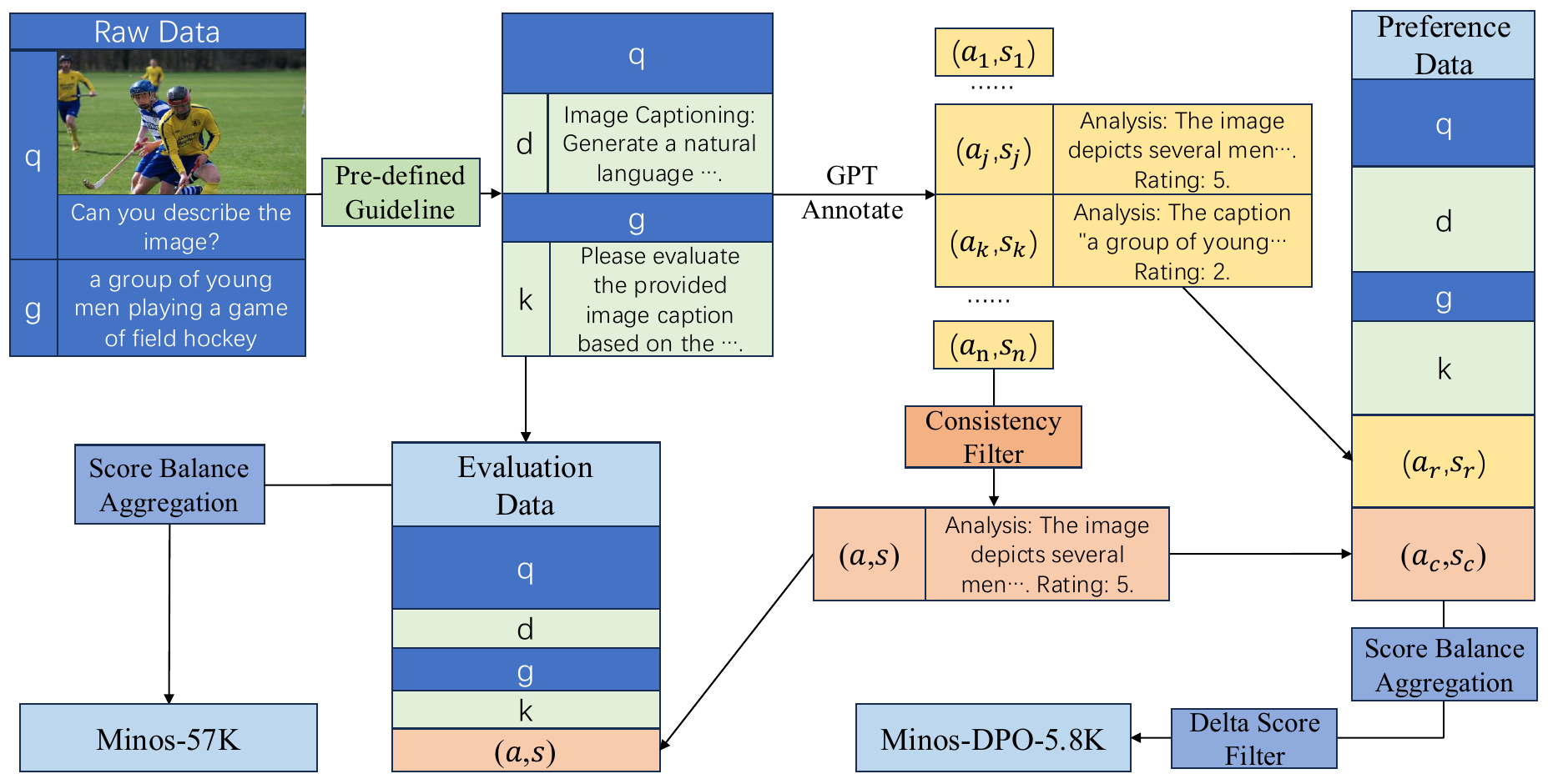}
    \caption{An overview of our dataset construction pipeline for \minosfinal and Minos-DPO-5.8K.}
    \label{fig:minos_framework}
\vspace{-10pt}
\end{figure*}

\section{Minos}
To develop multimodal evaluation model Minos, we first build structured evaluation inputs from raw multimodal task inputs and model responses with pre-defined evaluation guidelines, which provide task description and evaluation criterion, following our general multimodal evaluation data definition. We apply GPT-4o to generate evaluation output candidates and apply Consistency Filtering and Score Balancing to improve the quality of evaluation data, yielding the high-quality evaluation dataset \minosfinal. 
We then derive a preference alignment dataset by constructing evaluation preference pairs from GPT-annotated candidates, and perform Delta Score Filtering to produce preference dataset Minos-DPO-5.8K.
Minos is developed in two stages: SFT(supervised finetuning) on \minosfinal, and DPO alignment training on Minos-DPO-5.8K. Construction workflow is illustrated in the figure \ref{fig:minos_framework}.

\subsection{Data Definition}
We first introduce a more general definition of multimodal evaluation instance based on previous I2T evaluation data definition in LLaVA-Critic\cite{xiong2024llavacritic}.
Specifically, LLaVA-Critic define a evaluation instance as (Image, Question, Response, Reference, Evaluation Criterion, Score, Reason).
We generalize the image and question fields into a unified formulation consisting of a task description $d$, a task input query $q$, and a model output generation $g$, which allows for a consistent representation of evaluation data across both image-to-text (I2T) and text-to-image (T2I) tasks. Overall, a multimodal evaluation instance consists of a task description $d$, a task input query $q$, a model output generation $g$, an evaluation criterion $k$, and an optional reference answer $r$. The corresponding output of the evaluation instance includes an evaluation analysis $a$ and a 1–5 Likert scale pointwise evaluation score $s$. A single multimodal evaluation instance can be represented as: $(q, d, g, k, [r], a, s)$.
This standardized format provides a flexible and unified structure for representing evaluation of diverse multimodal generation tasks.
An specific example can be seen in the figure \ref{fig:data_instance} in appendix \ref{sec:examples}.

\subsection{Data Collection}
We start the data construction process by collecting raw multimodal task input query $q$ and its corresponding model output generation $g$. We first extract $<q, g>$ from existing high-quality human annotated evaluation dataset on both I2T and T2I tasks. We investigate three large-scale human-annotated evaluation datasets, including Image Captioning dataset Polaris\cite{wada2024polos}, Visual Question Answering dataset LAVE\cite{manas2024improving} and Text-to-Image Generation dataset ImageReward\cite{xu2023imagereward}. However, these evaluation data only contains scores without corresponding evaluation analyses which are an important component of the evaluation data, and many other tasks even lack corresponding human-annotated evaluation datasets. Therefore, we first extract some of the model responses from VLFeedback\cite{li2024vlfeedback} which contains responses across multiple multimodal tasks from different MLLMs, following previous settings\cite{xiong2024llavacritic}. We design the evaluation guidelines for different multimodal tasks to form evaluation inputs.
We then use GPT-4o to generate 10 evaluation output candidates which contain both analyses and scores for these evaluation inputs. We obtain 48k raw multi-response annotations for human-annotated evaluation datasets and 76k raw multi-response annotations on other tasks. In the end, we obtain \minosraw which contains 124k raw multi-response annotations, each annotation containing the corresponding evaluation input and a set of candidate evaluation outputs. Details about the raw dataset are provided in table \ref{table:all_data_tasks_details} in appendix \ref{appendix:data_details}.

\begin{table*}[htp]
    \centering
    \setlength{\tabcolsep}{1mm}{
    \renewcommand\arraystretch{1.2}
    \begin{tabular}{cccc}
    \toprule
    Evaluation Source & Multimodal Task             & Dataset                       & Data Size    \\\hline 
    \multirow{3}*{Human}             & Image Captioning            & Polaris                   & 3.2K         \\
                      & Visual Question Answering   & LAVE                              & 7.1K        \\
                      & Text-to-Image Generation    & ImageReward                       & 10.8K        \\\hline 
    \multirow{5}*{GPT-4o}               & Image Captioning            & SViT-D, LLaVA-D
                & 7.2K        \\
                      & Visual Question Answering   & LLaVAMed, LLaVA-C, comvint, SVIT-C   & 11.4k        \\ 
                      & Text Reading                & LLaVAR                            & 4.8k        \\ 
                      & Reasoning                   & LLaVA-R, SVIT-CR                       & 7.8k        \\ 
                      & Instruction Following       & PCAEVAL, M3IT, LRV-Instruction    & 5.2k         \\

    \bottomrule
    \end{tabular}}
    \caption{The Evaluation Source, Multimodal Task, Dataset, and corresponding Data Size of \minosfinal.  More details about the Dataset can be seen in Table \ref{table:all_data_tasks_details} in appendix \ref{appendix:data_details}.}
    \label{table:data_tasks_final}

\end{table*}

\subsection{Quality Control}
 Several existing multimodal evaluation models\cite{xiong2024llavacritic, wang2025unifiedrewardmodelmultimodal} focuses on increasing the variety and scale of evaluation data, while paying limited attention to rigorous evaluation data quality control. 
Based on prior research\cite{hu-etal-2024-themis}, we argue that evaluation performance depends more on data quality than on data scale, high-quality data Selection and Balance are critical for improving model evaluation capabilities. 

Therefore, we further impose both instance-level and dataset-level quality control strategies.
We try to obtain high quality evaluation data through Pre-defined Guideline, Consistency Filter and Score Balance, separately, as shown in figure \ref{fig:minos_framework}.


\noindent \textbf{Pre-defined Guideline}
Multimodal evaluation models are typically expected to assess a wide range of multimodal tasks, which may differ substantially from one another. Even for human evaluators, assessing different multimodal tasks typically requires specialized training to understand both the target task and the corresponding evaluation criterion which explains how to evaluate the model response. We therefore believe that, when evaluating a specific multimodal task instance, a multimodal evaluation model must understand both the target task and its associated evaluation criterion. Accordingly, we design evaluation guidelines which consists task descriptions and evaluation criterion for each of the six multimodal task categories in our dataset. Detailed guidelines are provided in the appendix \ref{sec:detailed_eval_criterion}.

\noindent \textbf{Consistency Filter}
To enhance the evaluation capability of Minos, we first design Consistency Filter method to obtain high-quality evaluation data over \minosraw. Specifically, for those raw evaluation multi-response annotations with human evaluation scores, we sample one evaluation result where the GPT-generated score matches the human score from 10 generated evaluation results. If none of the ten GPT scores align with the human judgment, the instance is discarded. For rest raw evaluation multi-response annotations that lack human-labeled evaluation scores, we first find the mode of the score in GPT-generated output candidates as the labeled evaluation score, and randomly select one evaluation analysis whose corresponding score matches the mode. We end up obtain 102k evaluation samples after Consistency Filter stage.

\noindent \textbf{Score Balance}
After applying the two preceding quality control procedures, we are able to ensure the quality of individual evaluation instances. However, when considered as a whole evaluation dataset, the dataset may still exhibit distributional imbalance. We first analyzed the overall score distribution of the combination of the consistency filtered dataset and observed a significant imbalance of evaluation scores as shown in Table \ref{table:score_distribution}. To address this issue, we manually balanced the score distribution by random removing some evaluation samples, resulting in the final evaluation dataset \minosfinal. The score distribution of the evaluation data can be seen in Table \ref{table:score_distribution}. More analysis about data can be seen in appendix \ref{appendix:data_details} and \ref{sec:human_study_res}. 


\begin{table*}[htp]
    \centering
    \small 
    \setlength{\tabcolsep}{0.5mm}{
    \renewcommand\arraystretch{1.5}
    \begin{tabular}{ccccccccccccccccccc}
    \toprule
    \multirow{2}*{Model}    & \multirow{2}*{Scale} &         \multicolumn{14}{c}{MLLM-as-a-Judge}  & \multirow{2}*{RichHF} & \multirow{2}*{GenAI} & All \\\cline{3-16} 
    
                            &       & CO      & C.C.      & Dif   & Graph    & Math    & Text   & WIT    & Chart & Vis   & CC    & M2W   & Sci   & Aes   & MM    &      &      & Ave. \\\hline 
    Gemini-2.5-Pro          & /     & 38.9    & 40.7      & 43.4  & 56.8     & 49.5    & 59.6   & 34.0   & 60.7  & 50.1  & 22.7  & -0.5  & 40.0  & 20.0  & 37.3  & 39.7 & 70.3 & 41.5 \\
    GPT-4o*                 & /     & 39.6    & 45.2      & 34.1  & 46.4     & 46.0    & 56.4   & 40.8   & 57.3  & 58.9  & 30.5  & 26.2  & 56.9  & 42.1  & 34.2  & 31.1 & 60.9 & \textbf{44.2} \\\hline 
    LLaVA-OV*               & 7B    & 22.4    & 2.4       & 6.30  & 18.9     & 9.70    & 26.5   & -13.5  & 27.4  & 22.7  & 8.10  & 3.0   & 26.1  & 24.9  & 26.2  & 5.85 & 16.4 & 14.6 \\ 
    LLaVA-OV*               & 72B   & 26.4    & 39.0      & 4.6   & 26.2     & 35.8    & 32.7   & 19.5   & 29.0  & 41.5  & 14.4  & 35.9  & 26.7  & 44.4  & 25.3  & 27.2 & 51.6 & 30.0 \\
    Qwen3-VL                & 8B    & 26.4    & 37.3      & 28.5  & 57.0     & 49.0    & 51.2   & 43.2   & 56.4  & 43.3  & 27.2  & -0.1  & 39.6  & 24.8  & 29.3  & 38.9 & 61.6 & 38.4 \\
    Prometheus-V*           & 7B    & 28.9    & 34.2      & 10.6  & 17.2     & 18.2    & 21.4   & 20.9   & 22.4  & 22.6  & 22.8  & 8.90  & 17.4  & 36.8  & 15.7  & 8.19 & 18.6 & 20.3 \\ 
    LLaVA-Critic*           & 7B    & 38.2    & 45.0      & 10.3  & 31.6     & 35.6    & 37.8   & 17.9   & 42.1  & 32.2  & 24.6  & 30.1  & 26.9  & 39.5  & 27.2  & 18.4 & 33.0 & 30.7 \\
    LLaVA-Critic*           & 72B   & 33.3    & 46.3      & 14.6  & 45.2     & 47.4    & 55.9   & 39.6   & 54.5  & 48.8  & 27.3  & 25.9  & 33.4  & 40.3  & 37.4  & 33.0 & 53.2 & 39.8 \\
    UnifiedReward\_L        & 7B    & 25.0    & 37.8      & 19.6  & 34.4     & 40.5    & 44.0   & 22.6   & 39.5  & 35.9  & 22.9  & 22.0  & 36.9  & 32.4  & 21.8  & 39.9 & 62.6 & 33.6 \\
    UnifiedReward\_Q        & 8B    & 29.3    & 35.1      & 22.3  & 44.0     & 46.6    & 43.4   & 29.0   & 55.3  & 37.9  & 25.0  & 17.2  & 43.1  & 30.6  & 33.6  & 40.4 & 62.7 & 37.2 \\
    Minos                   & 8B    & 32.8    & 41.1      & 25.1  & 54.1     & 51.3    & 58.3   & 41.4   & 62.1  & 47.1  & 24.3  & 23.6  & 50.7  & 28.7  & 39.8  & 36.0 & 60.2 & \textbf{42.3}  \\
    \bottomrule
    \end{tabular}}
    \caption{Main Result of Minos and other evaluation models on MLLM-as-a-Judge, RichHF-18K and GenAI-Bench. We present the \textbf{pearson-r} between the evaluation scores of models and the evaluation scores of human. We report results across two model categories: closed-source models, open-source MLLMs. We include results (which are referred as *) from previous researches\cite{chen2024mllm, xiong2024llavacritic}, and additionally evaluate all models on the text-to-image evaluation dataset RichHF-18K and GenAI-Bench. In each categories, we highlight in bold the model that achieves the highest consistency with human evaluations.}
\label{table:main_res}
\end{table*}

\subsection{Minos Training}
Firstly, we perform supervised-finetuning (SFT) using \minosfinal to construct an MLLM with basic evaluation capabilities.
Further, for MLLM development, it is common to follow the SFT stage with an alignment phase, which often leads to further performance improvements. We believe that this observation also holds for training multimodal evaluation models. Therefore, we introduce DPO Alignment stage after SFT stage to further improve the evaluation ability of Minos. 

To perform DPO alignment, we first obtain preference pairs from \minosraw. After applying the same quality control strategies, for each instance in the evaluation dataset, we apply the previously selected annotation through Consistency Filter as the chosen data. Among the remaining evaluation output candidates for the same sample, we identify the one with the largest score discrepancy from the selected sample and treat it as the rejected evaluation data to form a comparison pair. We discard the instance, where all candidate evaluation scores are identical to the previously selected good evaluation data. We obtain 38k evaluation preference pairs in the end. 

Unlike other tasks, the evaluation data contains evaluation scores. This enables us to quantify the strength of preference by comparing the score difference between the chosen and rejected instances within each preference pair. We compute the score gap between the chosen and rejected instances, and retain only pairs with a score difference no less than 2. The resulting 5.8K high-confidence preference pairs are used as the final dataset Minos-DPO-5.8K for DPO Alignment. Through further DPO alignment on Minos-DPO-5.8K after SFT on \minosfinal, we obtain our multimodal evaluation model Minos. More training details can be seen in appendix \ref{sec:training_detail}.

\section{Experiments}
\subsection{Experimental Setup}
                                                                                                      
\paragraph{Benchmark} Following the evaluation protocol established in LLaVA-Critic\cite{xiong2024llavacritic}, we adopt MLLM-as-a-Judge\cite{chen2024mllm} to assess the generalization performance of our evaluation model on various out-of-domain test datasets. MLLM-as-a-Judge consists of 5k evaluation samples spanning 14 datasets. More experimental details can be seen in appendix \ref{sec:exp_details}. 
Since MLLM-as-a-Judge doesn't contain T2I generation task, we additionally select two extra text-to-image generation evaluation dataset RichHF-18K\cite{liang2024richhumanfeedbacktexttoimage} and GenAI-Bench\cite{jiang2024genai} as test data for text-to-image generation task. More details about data integrity can be seen in appendix \ref{sec:data_integrity}.
Following previous setting\cite{xiong2024llavacritic}, we apply Pearson-r to measure the consistency between the model’s evaluation scores and human evaluation scores. We also show the results with correlation coefficient Kendall's Tau, details can be seen in appendix \ref{sec:kendall}.

\paragraph{Baselines}
We selected Gemini-2.5-Pro\cite{comanici2025gemini25pushingfrontier} and GPT-4o\cite{openai2024gpt4technicalreport} to evaluate the performance of closed-source models. For open-source models, we evaluated base MLLMs of different scales, including LLaVA-OV-7B, LLaVA-OV-72B\cite{li2024llavaonevisioneasyvisualtask} and Qwen3-VL-8B\cite{bai2025qwen3vltechnicalreport}, as well as prior multimodal evaluation models such as Prometheus-V\cite{lee2024prometheus} and LLaVA-Critic\cite{xiong2024llavacritic}. Specifically, Prometheus-V is built upon LLaVA-v1.5-7B\cite{liu2023llava}, while LLaVA-Critic includes both versions based on a 7B and a 72B LLaVA-OV. Recent work has introduced several multimodal reward models such as UnifiedReward\cite{wang2025unifiedrewardmodelmultimodal}, but most of them only support pairwise evaluation. We select UnifiedReward\cite{wang2025unifiedrewardmodelmultimodal}, which natively supports pointwise scoring, as a representative reward model for testing. We evaluate both the earlier version of the UnifiedReward presented in its paper, which is built on LLaVA-OV-7B(UnifiedReward\_L), as well as the recently released version based on Qwen3-VL-8B(UnifiedReward\_Q).

\subsection{Main Results}
The main results of Minos with other models on different multimodal tasks can be seen in Table \ref{table:main_res}. We evaluate our model against closed-source models, open-source MLLMs built on different backbones. As shown in the results, our model obtains the highest performance on the average of all multimodal generation tasks and achieves the highest agreement with human evaluations across 16 diverse multimodal evaluation tasks among open-source models. Compared with closed-source models, Minos achieves higher average evaluation consistency than Gemini-2.5-Pro, only performing slightly below GPT-4o with an average gap of 1.9 pearson-r on all multimodal generation tasks. In summary, Minos achieve state of the art on average evaluation consistency among open-source evaluation models, outperforming all existing open-source models. Compared to prior SoTA of open-source evaluation model LLaVA-Critic(72B), Minos achieves an average improvement of 2.5 Pearson-r on all multimodal generation tasks with smaller backbone. Compared to prior SoTA of open-source evaluation model UnifiedReward\_Q with the same scale, Minos achieves an average improvement of 5.1 Pearson-r on all multimodal generation tasks. More results of T2I specialized evaluation models on T2I evaluation dataset can be seen in table \ref{table:T2I_evaluators_results} in appendix \ref{sec:T2I evaluators}.

\subsection{Ablation Study}
\noindent \textbf{Ablation Study of Quality Control}
We conduct an ablation study to analyze the impact of quality control on the training data, and the results are presented in the table \ref{table:data_ablation_analysis}. 
Starting from the \minosraw dataset, we first randomly sample one output from 10 evaluation output candidates to construct each evaluation sample, without applying any explicit evaluation criterion and task description when building the evaluation sample. We use such samples to simulate and evaluate a low-control setting, where the evaluation data lacks predefined evaluation guidelines, Consistency Filtering, and Score Balancing constraints. This corresponds to the configuration reported in the first row of the table \ref{table:data_ablation_analysis}. 

We sequentially obtain the \minosfinal by progressively enabling the three modules of our quality control, Pre-defined Guideline, Consistency Filter, and Score Balancing. After full quality control is applied, the resulting dataset contains less than half of Minos-RAW and the data used in prior work, which is 113K in LLaVA-Critic\cite{xiong2024llavacritic} and 236K in UnifiedReward\cite{wang2025unifiedrewardmodelmultimodal}. However, the experimental findings indicate that the quality of evaluation data is more critical than its scale. Minos trained on the 57k quality-controlled evaluation samples exhibits stronger evaluation capability than Minos trained on the larger, unfiltered raw dataset. As more quality control methods are applied, the dataset scale generally decreases, yet the model’s average evaluation capability consistently improves across multiple multimodal tasks. More importantly, our experiments show that training an evaluation model on large-scale but low-quality evaluation data may harm its original evaluation capability, potentially making it underperform the base model prior to evaluation model training. This finding further highlights the critical importance of evaluation data quality.

\begin{table}[htp]
    \centering
    \setlength{\tabcolsep}{0.5mm}{
    \renewcommand\arraystretch{1.1}
    \begin{tabular}{ccccc}
    \toprule
    Score          & Consistency             & Evaluation    &  SFT Data             & All                \\
    Balance        & Filter                  & Guideline      &  Size                 & Ave.               \\\hline 
    \ding{55}      & \ding{55}               & \ding{55}     & 124k                  & 36.3               \\
    \ding{55}      & \ding{55}               & \ding{51}     & 124k                  & 37.1               \\
    \ding{55}      & \ding{51}               & \ding{51}     & 102k                  & 39.0                \\
    \ding{51}      & \ding{51}               & \ding{51}     & 57k                   & \textbf{40.9}        \\
    \bottomrule
    \end{tabular}}
    \caption{Result of Minos with pearson-r on MLLM-as-a-Judge, RichHF-18K and GenAI-Bench when training with different multimodal evaluation Data during SFT stage. 
    }
\label{table:data_ablation_analysis}
\end{table}

\begin{table}[htp]
    \centering
    \setlength{\tabcolsep}{3mm}{
    \renewcommand\arraystretch{1.1}
    \begin{tabular}{cccc}
    \toprule
    \multirow{2}*{DPO} & Delta Score   & DPO Data        & All                \\
                       & Filter        & Size            & Ave.               \\\hline 
    \ding{55}          & \ding{55}     & 0k              & 40.9               \\
    \ding{51}          & \ding{55}     & 35k             & 40.1               \\
    \ding{51}          & \ding{51}     & 5.8k            & \textbf{42.3}      \\
    \bottomrule
    \end{tabular}}
    \caption{Result of Minos with pearson-r on MLLM-as-a-Judge, RichHF-18K and GenAI-Bench when training with different dpo settings. 0k means original SFT model.
    }
\label{table:dpo_ablation_analysis}
\end{table}

\noindent \textbf{Ablation Study of DPO Alignment}
When developing large language models, a common practice is to perform preference alignment training after SFT (Supervised Fine-Tuning), which typically leads to further improvements in model capability. However, prior multimodal evaluation models, such as LLaVA-Critic and UnifiedReward, are trained solely via supervised fine-tuning on multimodal evaluation data, without applying subsequent alignment training such as DPO (Direct Preference Optimization) on evaluation model itself.

Based on result of our ablation study, we identify a potential explanation: naively constructing preference pairs and simply applying DPO alignment does not always yield clear gains in evaluation capability. In some cases, applying DPO alignment may even degrade the evaluation model’s evaluation performance relative to the model before alignment training, such as 40.9 to 40.1 in table \ref{table:dpo_ablation_analysis}. 
However, this does not imply that preference alignment is a wrong approach for training multimodal evaluation models. Our experimental results demonstrate that when we leverage the evaluation score of evaluation data to further filter DPO preference pairs by the score gap between chosen and rejected outputs, we can obtain a smaller set of high-quality preference pairs that provide a stronger alignment signal. Training Minos using these fewer but higher-quality DPO pairs leads to additional gains in evaluation capability, improving the average evaluation performance from 40.9 to 42.3 in table \ref{table:dpo_ablation_analysis}.
The ablation results on DPO alignment further demonstrate that evaluation performance is more sensitive to data quality than data scale, which holds for both the evaluation data used in SFT and the preference pairs used in DPO alignment.
Full experimental results can be seen in appendix \ref{sec:full_result}.



\section{Analysis}

\subsection{Evaluation between I2T and T2I}
To analyze the impact of unifying I2T (Image-to-Text) and T2I (Text-to-Image) tasks, we conduct extra experiments on the T2I and I2T training data during the SFT stage, whose results are show in table \ref{table:data_task_analysis}. 
We conduct extra experiments with three configurations: training on T2I-only evaluation data, training on I2T-only evaluation data with the same scale (10K samples), and training on full I2T-only evaluation data. 0k means original base model Qwen3-vl-8B. All these subsets are sampled from \minosfinal. In addition to reporting the overall average evaluation performance(All Ave.), we further compute the specific average performance on I2T and T2I tasks separately, namely I2T Ave. and T2I Ave.

The experimental results reveal that training on T2I evaluation data alone may impair the model’s evaluation capability on I2T tasks, with average performance on I2T tasks dropping from 36.7 to 25.4. A similar negative transfer is observed in the opposite direction, where training on I2T data only reduces T2I evaluation performance from 50.3 to 46.1.
In contrast, when we perform joint training using both T2I and I2T evaluation data, the evaluation model exhibits mutual performance enhancement between the two task categories, suggesting that the training from T2I and I2T evaluation samples can be complementary rather than conflicting. Compared to the T2I-only training setting, adding 47K I2T evaluation samples yields a 1.8 improvement in average T2I evaluation capability. Likewise, compared to the 47k I2T-only training baseline, incorporating T2I evaluation data leads to a 4.2 improvement in average I2T evaluation performance.
These findings indicate that building evaluation MLLMs capable of unified assessment across both T2I and I2T tasks provides further benefits than developing category-specific evaluation models.

\begin{table}[htp]
    \centering
    \setlength{\tabcolsep}{2mm}{
    \renewcommand\arraystretch{1.1}
    \begin{tabular}{cccccccc}
    \toprule
    \multirow{2}*{I2T}     & \multirow{2}*{T2I} & SFT Data   & I2T   & T2I      & All                \\
                           &                    & Size       & Ave.  & Ave.     & Ave.               \\\hline 
    \ding{55}              & \ding{55}          & 0k         & 36.7  & 50.3     & 38.4               \\
    \ding{55}              & \ding{51}          & 10k        & 25.4  & 49.1     & 28.3               \\
    \ding{51}              & \ding{55}          & 10k        & 34.1  & 46.2     & 35.6               \\
    \ding{51}              & \ding{55}          & 47k        & 35.3  & 46.1     & 36.6               \\
    \ding{51}              & \ding{51}          & 57k        & \textbf{39.5}     & \textbf{50.9}         & \textbf{40.9}       \\
    \bottomrule
    \end{tabular}}
    \caption{Results of Minos with pearson-r on MLLM-as-a-Judge, RichHF-18K and GenAI-Bench when training with different multimodal evaluation task only during SFT stage after applying quality control strategies. 
    }
\label{table:data_task_analysis}
\end{table}

\subsection{Impact of Evaluation Analysis}
Many previous researches\cite{xiong2024llavacritic, wang2025unifiedrewardmodelmultimodal} treat rationale generation during evaluation as an auxiliary capability to justify the reasonableness of model evaluations. We conduct an additional experiment to analyze whether including evaluation analysis supervision in evaluation data affects the performace of the evaluation model’s final scoring behavior. As shown in the table \ref{table:reason_analysis}, evaluation analysis generation not only improves the interpretability of model evaluations, but also enhances the agreement between the model’s final scores and human judgments.

More analysis can be seen in appendix \ref{sec:failure_modes} and \ref{sec:perturbation}.

\begin{table}[htp]
    \centering
    \setlength{\tabcolsep}{1.5mm}{
    \renewcommand\arraystretch{1.1}
    \begin{tabular}{cc}
    \toprule
     \multirow{2}*{SFT Data}                         & All                \\
                                 & Ave.               \\\hline 
     w/o evaluation analysis                  & 38.8                   \\
     w evaluation analysis                & \textbf{40.9}               \\
    \bottomrule
    \end{tabular}}
    \caption{Result of Minos with pearson-r on MLLM-as-a-Judge, RichHF-18K and GenAI-Bench when supervised finetuning with/without evaluation analysis on \minosfinal. 
    }
\label{table:reason_analysis}
\end{table}


\section{Conclusion}
In this work, we first collect and construct a high-quality, general multimodal evaluation dataset: \minosfinal. \minosfinal comprises multimodal evaluation samples spanning 6 common tasks and 15 datasets, covering both image-to-text and text-to-image settings. Each sample is accompanied by evaluation analysis and score after strict quality control process. 
We further apply more strict filtering
to obtain the high-quality evaluation preference
dataset Minos-DPO-5.8K.
We build our multimodal evaluation model Minos with supervised finetuning on \minosfinal and dpo alignment on Minos-DPO-5.8K. 
Averaged across all benchmark tasks, Minos achieves state-of-the-art (SoTA) performance among all open-source evaluation models, and even outperforms several closed-source models. 
Extensive experimental results demonstrate that the quality, task diversity of evaluation data, preference alignment, and evaluation analysis supervision together contribute to improved evaluation capability in evaluation MLLMs.

\section*{Limitations}


Some early multimodal human evaluation datasets are no longer accessible due to broken links, and more human-annotated datasets are continuously being proposed. Our work represents a snapshot collection of the currently available multimodal human evaluation datasets. As the field progresses, we anticipate the emergence of larger and higher-quality human-labeled datasets, which can support more reliable evaluation results and enable more comprehensive experimental analysis. 
Considering the limitations of computational resources, we did not build our method upon extremely large foundation models such as Qwen3-VL-70B. Although larger backbone models may yield stronger performance, they typically demand substantially higher inference compute and longer evaluation latency, which can diminish their practical utility in real-world deployment scenarios.


\section*{Ethical Considerations}
We follow the correct usage of the data and models with the corresponding license. Since real-world tasks are much different from well-defined evaluation instance, we recommend conducting manual spot-checks of the model's outputs during deployment on real-world tasks to ensure the reliability of its evaluations. Additionally, Minos is primarily developed as an evaluation model. If it is instead used as a reward model to optimize other models, it is important to consider whether this may lead to issues of over-optimization.


\section*{Acknowledgements}
This work was supported by Beijing Natural Science Foundation (L253001), Key Laboratory of Science, Technology and Standard in Press Industry (Key Laboratory of Intelligent Press Media Technology) and National Engineering Research Center of New Electronic Publishing Technologies. We appreciate the anonymous reviewers for their helpful comments. Xiaojun Wan is the contact author.


\bibliography{acl_latex}

\appendix

\section{Experimental Details}
\label{sec:exp_details}
We present the full name of each task mentioned in Table \ref{table:main_res}. The full name remains the same as \cite{chen2024mllm}.

\begin{table}[htp]
    \centering
    \begin{tabular}{c|c}   
    \toprule
        Task Name (Short) & Task Name (Full) \\
        \midrule
        CO & MS COCO \\
        C.C. & Conceptual Captions \\
        Dif & DiffusionDB \\
        Graph & InfographicVOA\\
        Math & MathVista\\
        Text & TextVOA\\
        WIT & WIT\\
        Chart & ChartOA\\
        Vis & VisIT-Bench\\
        CC & CC-3M Concept-balanced\\
        M2W & Mind2Web\\
        Sci & ScienceOA\\
        Aes & AesBench\\
        MM & MMvet\\
    \bottomrule
    \end{tabular}
    \caption{The full name of each task.}
    \label{table:full_name}
\end{table}

\section{Dataset Details}
\label{appendix:data_details}
Here, we present the details of the dataset distribution before and after filtering in table \ref{table:all_data_tasks_details}. The score distribution is shown in table \ref{table:score_distribution}.

\begin{table*}[htp]
    \centering
    \setlength{\tabcolsep}{2.5mm}{
    \renewcommand\arraystretch{1.2}
    \begin{tabular}{cccc}
    \toprule
    Data Source                 & Task Type                                                         &Minos-RAW                  & Minos-57K    \\\hline 
    Polaris                 & Image Captioning                                             & 6.7k(5.4\%)         & 3.2k(5.6\%)   \\
    LAVE                    & Visual Question Answering                            & 13.7k(11.0\%) & 7.1k(12.3\%)   \\
    ImageReward             & Text-to-Image Generation                             & 27.4k(22.1\%) & 10.8k(18.8\%)  \\
    Human-Eval              & Mixed                                                                         & 47.8k(38.5\%)       & 21.1k(36.6\%)  \\\hline 
    SViT-detail             & Image Captioning                                    & 4.9k(3.9\%)         & 4.0k(6.9\%)   \\
    LLaVA-detail            & Image Captioning                                            & 5.8k(4.7\%)         & 3.2k(5.6\%)   \\
    LLaVAMed                & Visual Question Answering                                      & 5.7k(4.6\%)         & 2.7k(4.7\%)  \\ 
    LLaVA-conversation      & Visual Question Answering                                                       & 7.4k(6.0\%)         & 2.2k(3.8\%)  \\ 
    comvint                 & Visual Question Answering                                        & 7.7k(6.2\%)         & 3.3k(5.7\%)  \\ 
    SVIT-conversation       & Visual Question Answering                                                       & 7.6k(6.1\%)         & 3.2k(5.6\%)  \\ 
    LLaVAR                  & Text Reading                                                    & 13.5k(10.9\%) & 4.8k(8.3\%)  \\ 
    LLaVA-reasoning         & Reasoning                                                           & 7.7k(6.2\%)         & 4.3k(7.5\%)  \\ 
    SVIT-complex\_reasoning & Reasoning                                                                  & 7.6k(6.1\%)         & 3.5k(6.1\%)  \\ 
    PCAEVAL                 & instruction following                      & 0.8k(0.6\%) & 0.4k(0.7\%)  \\
    M3IT                    & instruction following                          & 1.5k(1.2\%) & 0.8k(1.4\%)  \\
    LRV-Instruction         & instruction following                                         & 6.3k(5.1\%)         & 4.0k(6.9\%)  \\
    GPT-Eval                & Mixed                                                                           & 76.5k(61.6\%)       & 36.5k(63.4\%) \\\hline 
    All     & Mixed                                                                          & 124.2k              & 57.6k \\
                        
    \bottomrule
    \end{tabular}}
    \caption{
    The Data Source, Task Type and corresponding Data Size of Minos-57K and Minos-RAW are shown in table. Minos-RAW contains a total of 124k evaluation instances constructed from diverse sources. In Table \ref{table:data_tasks_final}, SViT-D refers to SViT-detail, SVIT-C refers to SViT-conversation, SVIT-CR refers to SVIT-complex\_reasoning. LLaVA-D refers to LLaVA-detail, LLaVA-C refers to LLaVA-conversation, LLaVA-R refers to LLaVA-reasoning.
}
    \label{table:all_data_tasks_details}

\end{table*}

\begin{table}
    \centering
    \setlength{\tabcolsep}{1mm}
    \renewcommand\arraystretch{1.2}
    \begin{tabular}{ccccccc}
    \toprule
    Score                    & \#1    & \#2   & \#3   & \#4   & \#5   & All \\ \hline 
    \multirow{2}*{RAW*}      & 9.7k   & 15k   & 16k   & 19k   & 64k   & 124k    \\
                             & 8\%    & 12\%  & 13\%  & 16\%  & 51\%  & 100\%   \\\hline
    Consistency              & 9.2k   & 9.8k  & 12k   & 13k   & 57k   & 102k    \\
    Filtered                 & 9\%    & 10\%  & 12\%  & 13\%  & 56\%  & 100\% \\ 
    \bottomrule
    \multirow{2}*{Final}     & 9.2k   & 9.8k  & 12k   & 13k   & 14k   & 57k    \\
                             & 16\%   & 17\%  & 21\%  & 23\%  & 23\%  & 100\% \\ 
    \bottomrule
    \end{tabular}
    \caption{Score distribution of RAW* and Final dataset. We apply Data Selection and Balance method to filter the Corpus. RAW* means we random selection one of ten GPT output candidates in Minos-RAW. We calculated the number and corresponding proportion of data samples for each score.}
    \label{table:score_distribution}
\end{table}

\begin{table}[p]
    \centering
    \setlength{\tabcolsep}{2mm}{
    \renewcommand\arraystretch{1.1}
    \begin{tabular}{cccc}
    \toprule
    Metrics                      & Model                  & RichHF          & GenAI     \\\hline
    \multirow{2}*{pearson-r}     & HPS v2                 & 19.5            & 31.0      \\
                                 & PickScore              & 22.4            & 34.7      \\\hline 
    \multirow{2}*{kendall's tau} & HPS v2                 & 13.1            & 23.1       \\
                                 & PickScore              & 18.3            & 26.0       \\ 
    \bottomrule
    \end{tabular}}
    \caption{Result of specialized T2I evaluation models on T2I evaluation dataset RichHF-18K and GenAI-Bench. We present the pearson-r and kendall's tau between the evaluation scores of Minos and the evaluation scores of human.}
\label{table:T2I_evaluators_results}
\end{table}

\begin{table}
    \centering
    \setlength{\tabcolsep}{1mm}
    \renewcommand\arraystretch{1.2}
    \begin{tabular}{ccccccc}
    \toprule
    Gold Score               & \#1       & \#2       & \#3       & \#4     & \#5      \\ \hline 
    accuracy              & 17.1\%    & 27.7\%    & 57.9\%    & 50.5\%  & 4.4\%   \\
    \bottomrule
    \end{tabular}
    \caption{The scoring accuracy of Minos conditioned on different gold score levels.}
    \label{table:score_acc}
\end{table}

\section{Example of Evaluation Data}
\label{sec:examples}
\begin{figure*}[t]
    \centering
    \includegraphics[width=16cm]{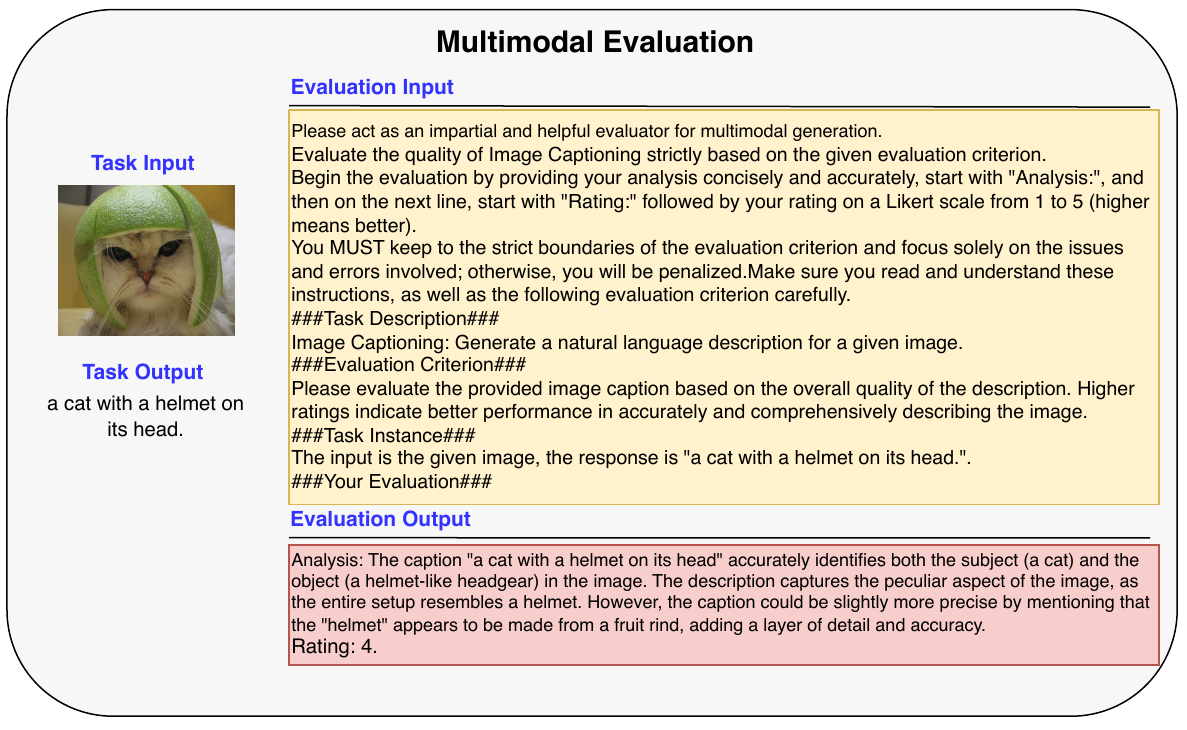}
    \caption{An example of Evaluation instance in \minosfinal. We select a sample constructed from a image captioning evaluation dataset Polaris to illustrate our data. Depending on the specific multimodal task, the task input can be an image, text, or a combination of both, while the task output can be either text or an image.}
    \label{fig:data_instance}
\vspace{-10pt}
\end{figure*}

Figure \ref{fig:data_instance} shows an example of Evaluation instance in \minosfinal.

\section{Training Details}
\label{sec:training_detail}
We build our model on top of Qwen3-VL-8B. During the SFT stage, we train the model for 2 epochs using a batch size of 64 and a learning rate of 1e-5. In the DPO stage, we train for 1 epoch with a learning rate of 2e-6, setting $\beta = 0.03$ and $\gamma = 0$. 
All other training configurations follow the default settings of Qwen3-vl-8B. We train the model with BF16 precision on 4 H100 GPUs. The SFT stage takes approximately 10 hours, while the DPO stage takes around 2 hours.

\section{Full Evaluation Results}
\label{sec:full_result}

We present full evaluation results in table \ref{table:full_result_data_ablation} (corresponding to table \ref{table:data_ablation_analysis}), table \ref{table:full_result_dpo} (corresponding to table \ref{table:dpo_ablation_analysis}), table \ref{table:full_result_I2T_T2I} (corresponding to table \ref{table:data_task_analysis}) and table \ref{table:full_result_reason} (corresponding to table \ref{table:reason_analysis}).

\begin{table*}[htp]
    \centering
    \small 
    \setlength{\tabcolsep}{0.6mm}{
    \renewcommand\arraystretch{1.5}
    \begin{tabular}{cccccccccccccccccc}
    \toprule
    \multirow{2}*{Setting}   &         \multicolumn{14}{c}{MLLM-as-a-Judge}                                                                               & \multirow{2}*{RichHF} & \multirow{2}*{GenAI} & All \\\cline{2-15} 
    
                             & CO      & C.C.      & Dif   & Graph    & Math    & Text   & WIT    & Chart & Vis   & CC    & M2W   & Sci   & Aes   & MM    &      &      & Ave.          \\\hline 
     $S_{1}(RAW)$            & 25.6    & 33.7      & 2.7   & 53.0     & 48.8    & 52.7   & 37.0   & 59.1  & 46.1  & 20.5  & 8.8   & 36.9  & 27.8  & 32.3  & 37.4 & 58.0 & 36.3          \\
     $S_{2}$                 & 25.5    & 40.1      & 17.1  & 53.4     & 47.7    & 50.0   & 43.3   & 58.7  & 44.6  & 17.4  & 11.5  & 27.6  & 23.4  & 33.2  & 38.8 & 60.9 & 37.1          \\ 
     $S_{3}$                 & 30.7    & 41.9      & 22.6  & 54.4     & 48.3    & 51.4   & 47.6   & 58.4  & 46.8  & 20.5  & 11.3  & 26.8  & 23.2  & 39.0  & 38.3 & 62.4 & 39.0          \\
     $S_{4}(Final)$          & 28.8    & 45.1      & 25.3  & 52.1     & 52.9    & 58.9   & 45.9   & 60.3  & 48.0  & 23.8  & 16.3  & 31.1  & 30.2  & 34.5  & 39.0 & 62.8 & \textbf{40.9} \\
    \bottomrule
    \end{tabular}}
    \caption{Detailed results of data analysis in table \ref{table:data_ablation_analysis}. $S_{i}$ represents the setting corresponding to the $i$-th row in table \ref{table:data_ablation_analysis}.}
\label{table:full_result_data_ablation}
\end{table*}

\begin{table*}[htp]
    \centering
    \small 
    \setlength{\tabcolsep}{0.6mm}{
    \renewcommand\arraystretch{1.5}
    \begin{tabular}{cccccccccccccccccc}
    \toprule
    \multirow{2}*{Setting}   &         \multicolumn{14}{c}{MLLM-as-a-Judge}                                                                               & \multirow{2}*{RichHF} & \multirow{2}*{GenAI} & All \\\cline{2-15} 
    
                             & CO      & C.C.      & Dif   & Graph    & Math    & Text   & WIT    & Chart & Vis   & CC    & M2W   & Sci   & Aes   & MM    &      &      & Ave.          \\\hline 
     $S_{1}$                 & 28.8    & 45.1      & 25.3  & 52.1     & 52.9    & 58.9   & 45.9   & 60.3  & 48.0  & 23.8  & 16.3  & 31.1  & 30.2  & 34.5  & 39.0 & 62.8 & 40.9      \\
     $S_{2}$                 & 24.8    & 39.9      & 24.7  & 52.0     & 52.0    & 54.1   & 35.9   & 61.0  & 40.0  & 28.1  & 18.3  & 46.1  & 33.0  & 38.5  & 36.8 & 57.0 & 40.1       \\
     $S_{3}$                 & 32.8    & 41.1      & 25.1  & 54.1     & 51.3    & 58.3   & 41.4   & 62.1  & 47.1  & 24.3  & 23.6  & 50.7  & 28.7  & 39.8  & 36.0 & 60.2 & \textbf{42.3}          \\ 
    \bottomrule
    \end{tabular}}
    \caption{Detailed results of data analysis in table \ref{table:dpo_ablation_analysis}. $S_{i}$ represents the setting corresponding to the $i$-th row in table \ref{table:dpo_ablation_analysis}.}
\label{table:full_result_dpo}
\end{table*}

\begin{table*}[htp]
    \centering
    \small 
    \setlength{\tabcolsep}{0.6mm}{
    \renewcommand\arraystretch{1.5}
    \begin{tabular}{cccccccccccccccccc}
    \toprule
    \multirow{2}*{Setting}   &         \multicolumn{14}{c}{MLLM-as-a-Judge}                                                                               & \multirow{2}*{RichHF} & \multirow{2}*{GenAI} & All \\\cline{2-15} 
    
                      & CO      & C.C.      & Dif   & Graph    & Math    & Text   & WIT    & Chart & Vis   & CC    & M2W   & Sci   & Aes   & MM    &      &      & Ave.          \\\hline 
     $S_{1}$          & 26.4    & 37.3      & 28.5  & 57.0     & 49.0    & 51.2   & 43.2   & 56.4  & 43.3  & 27.2  & -0.1  & 39.6  & 24.8  & 29.3  & 38.9 & 61.6 & 38.4         \\
     $S_{2}$          & 22.7    & 25.8      & 12.6  & 24.3     & 41.5    & 36.6   & 15.2   & 30.1  & 34.8  & 19.6  & 20.9  & 14.6  & 16.8  & 39.4  & 40.3 & 57.9 & 28.3         \\
     $S_{3}$          & 27.3    & 37.9      & 22.2  & 34.8     & 47.7    & 50.2   & 42.6   & 47.8  & 41.8  & 20.2  & 17.5  & 21.2  & 27.3  & 38.5  & 37.9 & 54.5 & 35.6         \\ 
     $S_{4}$          & 23.7    & 45.8      & 27.8  & 31.2     & 48.0    & 51.6   & 43.0   & 44.3  & 40.1  & 25.5  & 20.7  & 20.2  & 27.6  & 44.0  & 36.5 & 55.6 & 36.6         \\
     $S_{5}$          & 28.8    & 45.1      & 25.3  & 52.1     & 52.9    & 58.9   & 45.9   & 60.3  & 48.0  & 23.8  & 16.3  & 31.1  & 30.2  & 34.5  & 39.0 & 62.8 & \textbf{40.9} \\

    \bottomrule
    \end{tabular}}
    \caption{Detailed results of data analysis in table \ref{table:data_task_analysis}. $S_{i}$ represents the setting corresponding to the $i$-th row in table \ref{table:data_task_analysis}.}
\label{table:full_result_I2T_T2I}
\end{table*}

\begin{table*}[htp]
    \centering
    \small 
    \setlength{\tabcolsep}{0.6mm}{
    \renewcommand\arraystretch{1.5}
    \begin{tabular}{cccccccccccccccccc}
    \toprule
    \multirow{2}*{Setting}   &         \multicolumn{14}{c}{MLLM-as-a-Judge}                                                                               & \multirow{2}*{RichHF} & \multirow{2}*{GenAI} & All \\\cline{2-15} 
    
                             & CO      & C.C.      & Dif   & Graph    & Math    & Text   & WIT    & Chart & Vis   & CC    & M2W   & Sci   & Aes   & MM    &      &      & Ave.          \\\hline 
     $S_{1}$                 & 20.8    & 43.6      & 30.4  & 42.6     & 51.0    & 52.5   & 37.3   & 43.9  & 38.9  & 28.9  & 14.7  & 39.5  & 30.5  & 37.9  & 42.6 & 65.1 & 38.8          \\
     $S_{2}$                 & 28.8    & 45.1      & 25.3  & 52.1     & 52.9    & 58.9   & 45.9   & 60.3  & 48.0  & 23.8  & 16.3  & 31.1  & 30.2  & 34.5  & 39.0 & 62.8 & \textbf{40.9} \\ 
    \bottomrule
    \end{tabular}}
    \caption{Detailed results of data analysis in table \ref{table:reason_analysis}. $S_{i}$ represents the setting corresponding to the $i$-th row in table \ref{table:reason_analysis}.}
\label{table:full_result_reason}
\end{table*}

\section{Results with kendall's tau}
\label{sec:kendall}
We show the result of models in table \ref{table:main_res} with correlation coefficient kendall's tau. The results are shown in table \ref{table:main_res_kendall}. Since previous researches\cite{chen2024mllm, xiong2024llavacritic} don't include result with kendall's tau, we reproduce all models to obtain evaluation results with kendall's tau. We also present the pearson-r of our re-implemented models in table \ref{table:main_res_pearsonr_reimplement}. For rigor, the results compared in the main text are primarily cited from previously published papers whenever possible.

\begin{table*}[htp]
    \centering
    \small 
    \setlength{\tabcolsep}{0.5mm}{
    \renewcommand\arraystretch{1.5}
    \begin{tabular}{cccccccccccccccccc}
    \toprule
    \multirow{2}*{Model}   &         \multicolumn{14}{c}{MLLM-as-a-Judge}                                                                               & \multirow{2}*{RichHF} & \multirow{2}*{GenAI} & All \\\cline{2-15} 
    
                           &  CO    & C.C.       & Dif   & Graph    & Math    & Text   & WIT    & Chart & Vis   & CC     & M2W  & Sci   & Aes   & MM    &      &       & Ave. \\\hline 
    Gemini-2.5-Pro         &  29.9  & 34.1       & 36.0  & 48.1     & 42.1    & 47.4   & 28.6   & 51.2  & 40.5  & 18.9   & -0.2 & 34.5  & 18.8  & 23.2  & 33.9 & 61.6  & \textbf{34.3} \\
    GPT-4o                 &  23.1  & 28.8       & 16.1  & 36.8     & 35.9    & 39.6   & 30.3   & 47.2  & 31.9  & 18.3   & 1.7  & 28.8  & 20.4  & 29.0  & 27.1 & 53.6  & 29.3\\\hline 
    LLaVA-OV(7B)           &  15.3  & 16.9       & 10.4  & 12.9     & 11.5    & 19.8   & 3.3    & 28.2  & 19.6  & 15.9   & 17.5 & 6.6   & 26.9  & 26.0  & 3.15 & 13.6  & 15.5 \\ 
    LLaVA-OV(72B)          &  20.2  & 13.3       & 7.5   & 14.2     & 24.1    & 22.2   & 4.6    & 19.7  & 27.2  & 21.9   & 19.0 & 13.8  & 19.6  & 26.9  & 24.3 & 46.3  & 20.3 \\ 
    Qwen3-VL(8B)           &  17.3  & 31.2       & 23.6  & 49.1     & 40.9    & 41.0   & 37.9   & 49.3  & 36.4  & 22.0   & -0.7 & 31.7  & 22.9  & 22.6  & 33.7 & 53.6  & 32.0 \\
    Prometheus-V(7B)       &  11.2  & 9.10       & 9.40  & 12.7     & 11.1    & 17.8   & 18.0   & 8.40  & 11.5  & 15.5   & 11.4 & 8.20  & 24.1  & 11.0  & 6.55 & 16.0  & 12.6 \\ 
    LLaVA-Critic(7B)       &  19.2  & 31.6       & 10.3  & 30.1     & 23.3    & 29.1   & 8.3    & 25.0  & 25.4  & 15.6   & 14.8 & 7.7   & 28.6  & 25.3  & 16.8 & 29.8  & 21.3 \\ 
    LLaVA-Critic(72B)      &  30.3  & 33.8       & 2.9   & 34.8     & 38.1    & 37.4   & 13.5   & 43.1  & 34.8  & 20.4   & 7.6  & 25.6  & 17.0  & 29.3  & 28.9 & 46.8  & 27.8\\
    UnifiedReward\_L(7B)   &  23.7  & 31.9       & 13.8  & 29.5     & 34.2    & 35.0   & 20.2   & 35.7  & 30.0  & 18.3   & 12.5 & 31.4  & 27.0  & 14.9  & 33.4 & 54.4 & 27.9 \\
    UnifiedReward\_Q(8B)   &  26.9  & 30.3       & 17.6  & 37.7     & 37.8    & 35.4   & 29.5   & 48.7  & 29.2  & 21.4   & 7.9  & 34.8  & 20.9  & 28.3  & 33.6 & 54.7  & 30.9 \\
    Minos(8B)              &  24.9  & 35.7       & 23.1  & 48.5     & 41.4    & 44.5   & 36.6   & 50.9  & 35.0  & 25.2   & 12.0 & 38.7  & 23.8  & 25.8  & 31.5 & 55.9  & \textbf{34.6} \\
    \bottomrule
    \end{tabular}}
    \caption{Main Result of Minos and other evaluation models on MLLM-as-a-Judge, RichHF-18K and GenAI-Bench. We present the \textbf{kendall's tau} between the evaluation scores of models and the evaluation scores of human. We report results across two model categories: closed-source models, open-source MLLMs. Since previous researches\cite{chen2024mllm, xiong2024llavacritic} don't include result with kendall's tau, we reproduce all models to obtain evaluation results with kendall's tau. For models in each categories, we highlight in bold the model that achieves the highest consistency with human evaluations.}
\label{table:main_res_kendall}
\end{table*}

\begin{table*}[htp]
    \centering
    \small 
    \setlength{\tabcolsep}{0.5mm}{
    \renewcommand\arraystretch{1.5}
    \begin{tabular}{cccccccccccccccccc}
    \toprule
    \multirow{2}*{Model}   &         \multicolumn{14}{c}{MLLM-as-a-Judge}                                                                               & \multirow{2}*{RichHF} & \multirow{2}*{GenAI} & All \\\cline{2-15} 
    
                           &  CO    & C.C.       & Dif   & Graph    & Math    & Text   & WIT    & Chart & Vis   & CC     & M2W   & Sci   & Aes   & MM    &      &       & Ave. \\\hline 
    Gemini-2.5-Pro         &  38.9  & 40.7       & 43.4  & 56.8     & 49.5    & 59.6   & 34.0   & 60.7  & 50.1  & 22.7   & -0.5  & 40.0  & 20.0  & 37.3  & 39.7 & 70.3  & \textbf{41.5} \\
    GPT-4o                 &  30.1	& 33.0	     & 22.2	 & 44.3	    & 44.3	  & 52.8   & 32.8	& 57.4	& 40.1	& 22.4	 & 6.5	 & 36.2	 & 29.3	 & 31.2	 & 31.1	& 60.9	& 35.9 \\\hline 
    LLaVA-OV(7B)           &  20.6  & 19.4       & 13.6  & 21.1     & 14.8    & 28.2   & 4.9    & 30.3  & 29.6  & 16.9   & 34.2  & 13.5  & 42.3  & 39.1  & 5.85 & 16.4  & 21.9 \\ 
    LLaVA-OV(72B)          &  27.5  & 12.4       & 9.7   & 20.1     & 31.6    & 31.3   & 8.9    & 26.0  & 35.1  & 25.6   & 31.5  & 27.2  & 26.3  & 37.2  & 27.2 & 51.6  & 26.8 \\ 
    Qwen3-VL(8B)           &  26.4  & 37.3       & 28.5  & 57.0     & 49.0    & 51.2   & 43.2   & 56.4  & 43.3  & 27.2   & -0.1  & 39.6  & 24.8  & 29.3  & 38.9 & 61.6  & 38.4 \\ 
    Prometheus-V(7B)       &  13.4  & 10.3       & 12.2  & 15.4     & 13.7    & 21.3   & 19.7   & 13.5  & 14.2  & 16.1   & 18.1  & 11.9  & 36.3  & 15.8  & 8.19 & 18.6  & 16.2 \\ 
    LLaVA-Critic(7B)       &  25.9  & 34.3       & 15.1  & 37.9     & 28.5    & 38.4   & 17.6   & 32.0  & 34.1  & 20.1   & 23.1  & 15.6  & 39.2  & 32.3  & 18.4 & 33.0  & 27.8 \\ 
    LLaVA-Critic(72B)      &  37.0  & 36.0       & 8.6   & 43.1     & 45.7    & 46.8   & 20.7   & 53.0  & 44.0  & 26.5   & 14.7  & 35.2  & 27.1  & 42.2  & 33.0 & 53.2  & 35.4 \\
    UnifiedReward\_L(7B)   &  25.0  & 37.8       & 19.6  & 34.4     & 40.5    & 44.0   & 22.6   & 39.5  & 35.9  & 22.9   & 22.0  & 36.9  & 32.4  & 21.8  & 39.9 & 62.6  & 33.6 \\
    UnifiedReward\_Q(8B)   &  29.3  & 35.1       & 22.3  & 44.0     & 46.6    & 43.4   & 29.0   & 55.3  & 37.9  & 25.0   & 17.2  & 43.1  & 30.6  & 33.6  & 40.4 & 62.7  & 37.2 \\
    Minos(8B)              &  32.8  & 41.1       & 25.1  & 54.1     & 51.3    & 58.3   & 41.4   & 62.1  & 47.1  & 24.3   & 23.6  & 50.7  & 28.7  & 39.8  & 36.0 & 60.2  & \textbf{42.3} \\
    \bottomrule
    \end{tabular}}
    \caption{Main Result of Minos and other evaluation models on MLLM-as-a-Judge, RichHF-18K and GenAI-Bench. We present the \textbf{pearson-r} between the evaluation scores of models and the evaluation scores of human. We report results across two model categories: closed-source models, open-source MLLMs. We reproduce all models to obtain evaluation results with pearson-r. For models in each categories, we highlight in bold the model that achieves the highest consistency with human evaluations.}
\label{table:main_res_pearsonr_reimplement}
\end{table*}


\section{More Results with T2I evaluators}
\label{sec:T2I evaluators}
To enable a more comprehensive comparison among multimodal evaluation models, we additionally benchmarked specialized text-to-image (T2I) evaluation models, such as HPS v2\cite{wu2023human} and PickScore\cite{kirstain2023pickapicopendatasetuser}, on out-of-domain T2I evaluation datasets RichHF-18K and GenAI-Bench.
As shown in table \ref{table:T2I_evaluators_results}, even a specialized T2I evaluation model does not surpass Minos in evaluation capability on T2I tasks, which further demonstrates the advantage of jointly training text-to-image and image-to-text evaluation abilities.

\section{Detailed Evaluation Guidelines}
\label{sec:detailed_eval_criterion}
We present the evaluation guidelines for six distinct tasks, as shown in the figure \ref{fig:Image Caption Evaluation Guideline}, \ref{fig:Visual Question Answering Evaluation Guideline}, \ref{fig:Image Generation Evaluation Guideline}, \ref{fig:Text Reading Evaluation Guideline}, \ref{fig:Reasoning Evaluation Guideline} and \ref{fig:Instruction Following Evaluation Guideline}.

\section{Human Study}
\label{sec:human_study_res}
To validate the effectiveness of our filtering pipeline, we randomly sampled 50 evaluation instances from the filtered GPT-generated data and conducted human assessment. First, we measured the agreement between GPT scores and human scores on this subset, obtaining a consistency of 63.9, indicating that our filtering strategy successfully retains samples with relatively reliable judgments.

We further evaluated the quality of the generated rationales along three dimensions: Factual Accuracy, Relevance to the Evaluation Criterion, and Coherence. Under a 1–5 scoring scale, the average scores are 4.62, 4.64, and 4.56, respectively. These results demonstrate that the filtered synthetic data maintains high-quality rationales in addition to reasonably aligned scoring. The evaluation guidelines can be seen in figure \ref{fig:Human Evaluation Guideline of Rationale}. Annotators were informed about the purpose of the task and consented to perform the annotation. The task did not involve sensitive or personal data.

\section{Analysis of Failure Modes}
\label{sec:failure_modes}
\begin{figure*}[t]
    \centering
    \includegraphics[width=16cm]{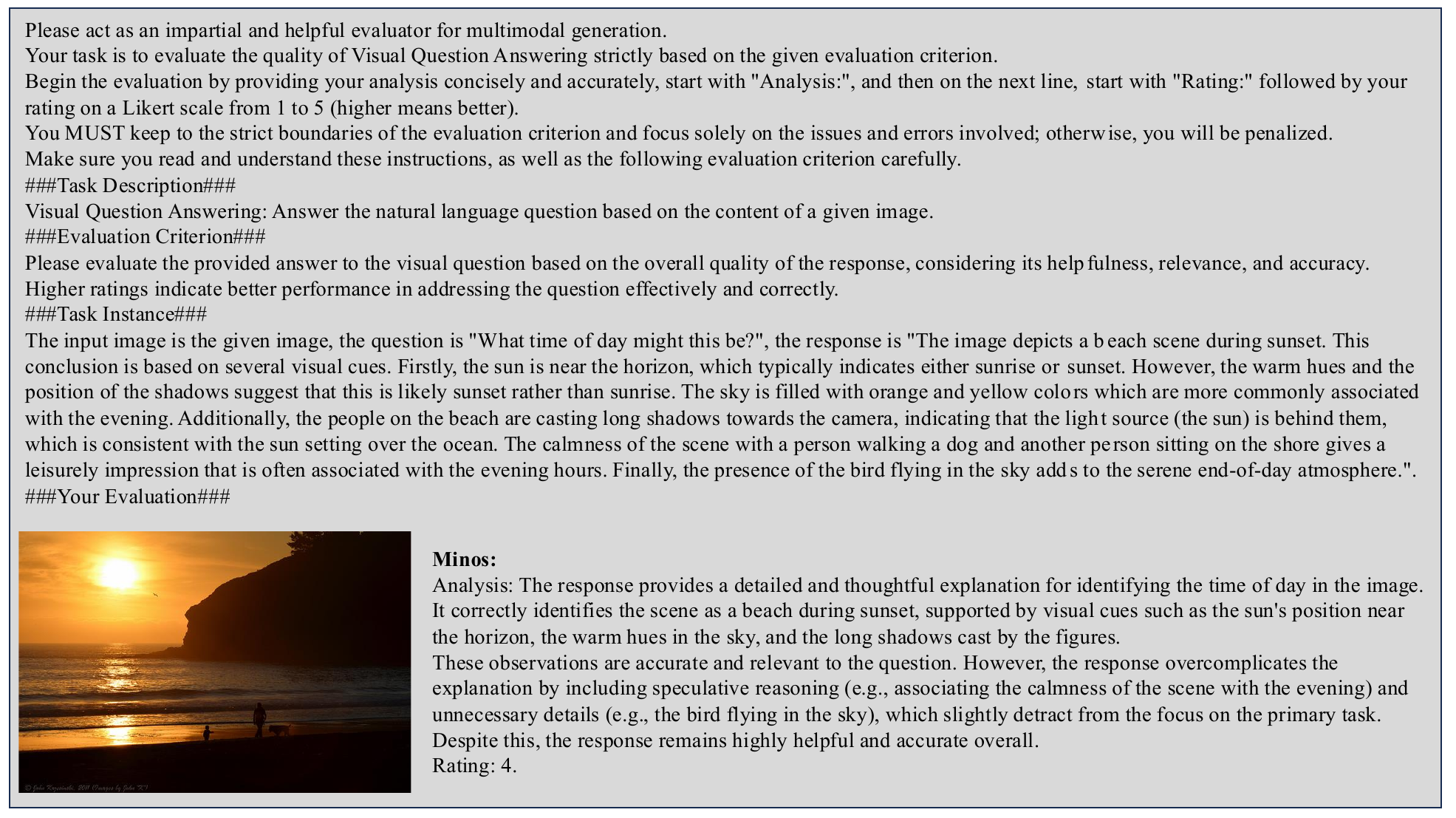}
    \caption{Score discrepancy in an example from MLLM-as-a-Judge: Human Evaluation (5) compared with Minos Evaluation (4).}
    \label{fig:gold_score_5_fail}
\vspace{-10pt}
\end{figure*}

We further analyze Minos’s scoring accuracy conditioned on different gold score levels. The detailed results are shown in the table \ref{table:score_acc}.
It can be observed that Minos achieves relatively high accuracy on samples with gold scores of 3 and 4, while its accuracy is lower for scores of 1 and 5, particularly for score 5.
Through case studies, we find that Minos tends to be conservative when assigning a model response corresponding to perfect score 5, especially for complex tasks. It is often reluctant to give full marks even when the output is of high quality, which may explain a substantial portion of the observed errors and point out promising direction for developing better multimodal evaluation models. We show an example of this failure from MLLM-as-a-Judge in figure \ref{fig:gold_score_5_fail}.

\begin{figure*}[t]
    \centering
    \includegraphics[width=16cm]{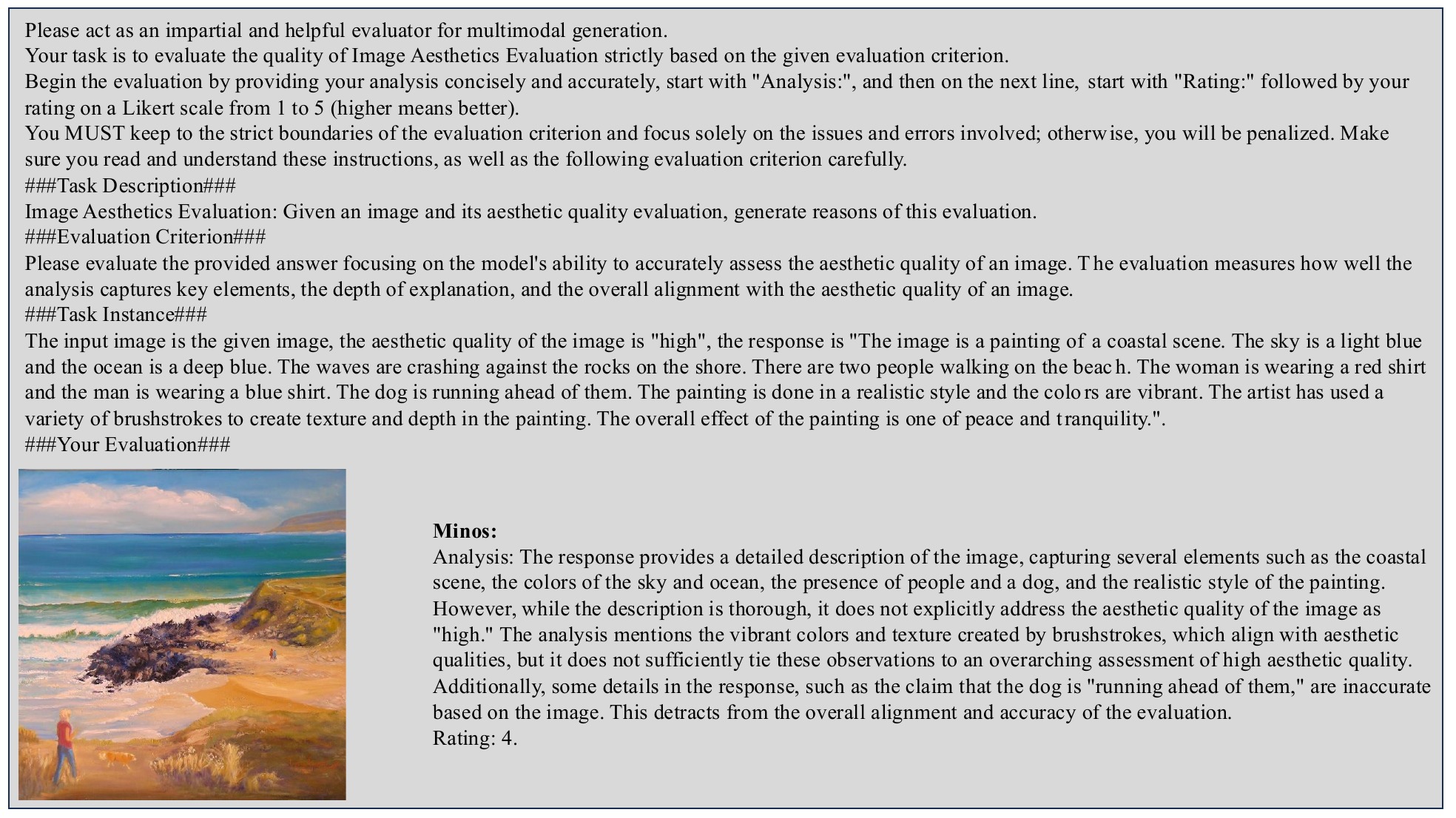}
    \caption{Score discrepancy in an example from AesBench: Human Evaluation (2) compared with Minos Evaluation (4).}
    \label{fig:AesBench_Failure}
\vspace{-10pt}
\end{figure*}

Meanwhile, we observe that Minos exhibits relatively low evaluation consistency on certain tasks, such as AesBench. We select a failure case for analysis, as illustrated in the figure \ref{fig:AesBench_Failure}.
In the example, Minos correctly identifies some inaccuracies in the model’s response; however, it fails to point out other aspects that cannot be reliably inferred, such as the description about the “man,” which is not clearly supported by the image. Moreover, even when the model’s response contains partially incorrect analysis, Minos still assigns a score of 4. This suggests that, in certain instances, the identification of errors does not substantially affect its final scoring. Such behavior is inconsistent with the evaluation criterion of “how well the analysis captures key elements.” A possible explanation is that the task itself involves evaluating the model’s evaluation of image aesthetic quality, which may introduce additional complexity and challenge compared to other non-evaluation tasks.

\section{Data Integrity}
\label{sec:data_integrity}
The definition of an evaluation sample follows Section 3.1: (q, d, g, k, [r], a, s). Our training data consists of human-annotated evaluation datasets and GPT-4o-annotated data. Since all references in the test sets are human scores, we only need to verify whether the human-annotated evaluation data in training overlaps with the human-annotated data in testing. The human scores (s) in both training and test sets are independently annotated in their respective original papers. 

We further detail the data sources. MLLM\_Judge\cite{chen2024mllm}: MLLM\_Judge generates model responses mostly using LLaVA-1.5-
13b, LLaVA-1.6-34b\cite{liu2023visual}, CogVLM\cite{wang2024cogvlm}, GPT-4V\cite{2023GPT4VisionSC}, Qwen-VL-Max\cite{bai2023qwenvlversatilevisionlanguagemodel} and Gemini\cite{team2023gemini}. Therefore, many evaluation instance in our human-annotated evaluation data of training dataset differs from this test data in terms of model response (g) which comes from different applied models.

Polaris\cite{wada2024polos}: Polaris uses MSCOCO\cite{lin2014microsoft} and NoCaps\cite{Agrawal_2019} and generates model responses with SAT\cite{xu2015show}, $M^2$-Transformer\cite{cornia2020meshed}, VinVL\cite{zhang2021vinvl}, GRIT\cite{10.1007/978-3-031-20059-5_10}, BLIP\cite{li2022blip}, and OFA\cite{wang2022ofa} for captioning tasks. While MLLM\_Judge also contains data derived from MSCOCO, those correspond to VQA tasks rather than Polaris’s captioning setting. Moreover, the models used to generate responses are different, resulting in different (g), (a), and (s).

LAVE\cite{manas2024improving}: LAVE collects human judgments for answers generated by BLIP-2\cite{li2023blip}, PromptCap\cite{hu2022promptcap} and BLIP\cite{li2022blip}. Since most of the underlying generation models differ, the corresponding responses (g), annotations (a), and scores (s) are also different.

ImageReward\cite{xu2023imagereward}: ImageReward uses real user prompts from DiffusionDB\cite{wang2023diffusiondb}. In contrast, our test set RichHF\cite{liang2024richhumanfeedbacktexttoimage} is derived from the Pick-a-Pic\cite{kirstain2023pick} dataset, annotated via the Pick-a-Pic web application. Additionally, GenAI-Bench\cite{jiang2024genai} consists of 1,600 challenging real-world prompts sourced from professional designers. These data sources are distinct from ImageReward, and thus do not overlap. Althought Dif dataset in MLLM\_Judge apply data from DiffusionDB, it is used for I2T task which is different from T2I task in ImageReward.

\section{Perturbation Experiments}
\label{sec:perturbation}
We additionally conduct a common multimodal adversarial experiment. Specifically, we randomly replace the input images to construct corrupted samples and evaluate Minos under this setting.
After random image replacement, Minos’s overall scoring pearson-r consistency drops sharply to 8.5. This substantial degradation indicates that Minos is highly sensitive to visual information and does not rely solely on textual cues. Instead, it performs evaluation by jointly considering both the input image and the text, demonstrating genuine multimodal evaluation behavior.

We further conduct a perturbation experiment on the MLLM\_Judge test set. Specifically, we randomly replace 25\% of the spans in the evaluated model responses to construct corrupted versions.
We then compare the scores assigned by Minos before and after perturbation. The results show that 71.27\% of the perturbed responses receive lower scores by Minos than their original counterparts.
This indicates that Minos is able to distinguish between higher and lower quality responses and is sensitive to content degradation, rather than assigning scores arbitrarily.

\begin{figure*}[t]
        \centering
        \begin{tcolorbox}[title=Image Captioning Evaluation Guideline]
        Please act as an impartial and helpful evaluator for multimodal generation.
        Evaluate the quality of Image Captioning strictly based on the given evaluation criterion.\par\vspace{\baselineskip}
        Begin the evaluation by providing your analysis concisely and accurately, start with "Analysis:", and then on the next line, start with "Rating:" followed by your rating on a Likert scale from 1 to 5 (higher means better).
        You MUST keep to the strict boundaries of the evaluation criterion and focus solely on the issues and errors involved; otherwise, you will be penalized.
        Make sure you read and understand these instructions, as well as the following evaluation criterion carefully.\par\vspace{\baselineskip}

        \#\#\#Task Description\#\#\#
        
        Image Captioning: Generate a natural language description for a given image.\par\vspace{\baselineskip}

        \#\#\#Evaluation Criterion\#\#\#

        Please evaluate the provided image caption based on the overall quality of the description. Higher ratings indicate better performance in accurately and comprehensively describing the image.\par\vspace{\baselineskip}

        \#\#\#Task Instance\#\#\#

        The input is the given image, the response is \{model response\}.\par\vspace{\baselineskip}

        \#\#\#Your Evaluation\#\#\#

        \end{tcolorbox}
\caption{Image Captioning Evaluation Guideline.}
\label{fig:Image Caption Evaluation Guideline}
\end{figure*}

\begin{figure*}[htp]
        \centering
        \begin{tcolorbox}[title=Visual Question Answering Evaluation Guideline]
        Please act as an impartial and helpful evaluator for multimodal generation.
        Evaluate the quality of Visual Question Answering strictly based on the given evaluation criterion.\par\vspace{\baselineskip}
        Begin the evaluation by providing your analysis concisely and accurately, start with "Analysis:", and then on the next line, start with "Rating:" followed by your rating on a Likert scale from 1 to 5 (higher means better).
        You MUST keep to the strict boundaries of the evaluation criterion and focus solely on the issues and errors involved; otherwise, you will be penalized.
        Make sure you read and understand these instructions, as well as the following evaluation criterion carefully.\par\vspace{\baselineskip}

        \#\#\#Task Description\#\#\#
        
        Visual Question Answering: Answer the natural language question based on the content of a given image.\par\vspace{\baselineskip}

        \#\#\#Evaluation Criterion\#\#\#

        Please evaluate the provided answer to the visual question based on the overall quality of the response, considering its helpfulness, relevance, and accuracy. Higher ratings indicate better performance in addressing the question effectively and correctly.\par\vspace{\baselineskip}

        \#\#\#Task Instance\#\#\#

        The input image is the given image, the question is \{VQA question\}, the response is \{model response\}.\par\vspace{\baselineskip}

        \#\#\#Your Evaluation\#\#\#

        \end{tcolorbox}
\caption{Visual Question Answering Evaluation Guideline.}
\label{fig:Visual Question Answering Evaluation Guideline}
\end{figure*}

\begin{figure*}[htp]
        \centering
        \begin{tcolorbox}[title=Image Generation Evaluation Guideline]
        Please act as an impartial and helpful evaluator for multimodal generation.
        Evaluate the quality of Image Generation strictly based on the given evaluation criterion.\par\vspace{\baselineskip}
        Begin the evaluation by providing your analysis concisely and accurately, start with "Analysis:", and then on the next line, start with "Rating:" followed by your rating on a Likert scale from 1 to 5 (higher means better).
        You MUST keep to the strict boundaries of the evaluation criterion and focus solely on the issues and errors involved; otherwise, you will be penalized.
        Make sure you read and understand these instructions, as well as the following evaluation criterion carefully.\par\vspace{\baselineskip}

        \#\#\#Task Description\#\#\#
        
        Image Generation: Generate the image following the input text.\par\vspace{\baselineskip}

        \#\#\#Evaluation Criterion\#\#\#

        Please evaluate the generated image based on how well the generated image matches the given text description and the overall quality of the image.\par\vspace{\baselineskip}

        \#\#\#Task Instance\#\#\#

        The input text is \{input text\}, the generated response image is the given image.\par\vspace{\baselineskip}

        \#\#\#Your Evaluation\#\#\#

        \end{tcolorbox}
\caption{Image Generation Evaluation Guideline.}
\label{fig:Image Generation Evaluation Guideline}
\end{figure*}

\begin{figure*}[htp]
        \centering
        \begin{tcolorbox}[title=Text Reading Evaluation Guideline]
        Please act as an impartial and helpful evaluator for multimodal generation.
        Evaluate the quality of Text Reading strictly based on the given evaluation criterion.\par\vspace{\baselineskip}
        Begin the evaluation by providing your analysis concisely and accurately, start with "Analysis:", and then on the next line, start with "Rating:" followed by your rating on a Likert scale from 1 to 5 (higher means better).
        You MUST keep to the strict boundaries of the evaluation criterion and focus solely on the issues and errors involved; otherwise, you will be penalized.
        Make sure you read and understand these instructions, as well as the following evaluation criterion carefully.\par\vspace{\baselineskip}

        \#\#\#Task Description\#\#\#
        
        Text Reading: Given an image containing text information and a question related to text information in the image, analyze this figure in detail and answer the question.\par\vspace{\baselineskip}

        \#\#\#Evaluation Criterion\#\#\#

        Please evaluate the provided answer to the visual question based on the overall quality of the response, considering its helpfulness, relevance, and accuracy. Higher ratings indicate better performance in addressing the question effectively and correctly.\par\vspace{\baselineskip}

        \#\#\#Task Instance\#\#\#

        The input image is the given image, the question is \{input question\}, the response is \{model response\}.\par\vspace{\baselineskip}

        \#\#\#Your Evaluation\#\#\#

        \end{tcolorbox}
\caption{Text Reading Evaluation Guideline.}
\label{fig:Text Reading Evaluation Guideline}
\end{figure*}


\begin{figure*}[htp]
        \centering
        \begin{tcolorbox}[title=Reasoning Evaluation Guideline]
        Please act as an impartial and helpful evaluator for multimodal generation.
        Evaluate the quality of Reasoning strictly based on the given evaluation criterion.\par\vspace{\baselineskip}
        Begin the evaluation by providing your analysis concisely and accurately, start with "Analysis:", and then on the next line, start with "Rating:" followed by your rating on a Likert scale from 1 to 5 (higher means better).
        You MUST keep to the strict boundaries of the evaluation criterion and focus solely on the issues and errors involved; otherwise, you will be penalized.
        Make sure you read and understand these instructions, as well as the following evaluation criterion carefully.\par\vspace{\baselineskip}

        \#\#\#Task Description\#\#\#
        
        Reasoning: Given an image with a question, generate a detailed analysis with the answer of the question.\par\vspace{\baselineskip}

        \#\#\#Evaluation Criterion\#\#\#

        Please evaluate the provided answer focusing on the alignment and coherence of the model's reasoning concerning the image. The evaluation measures the extent to which the explanation responds effectively to the question, accurately represents the image's content, and provides a useful answer.\par\vspace{\baselineskip}

        \#\#\#Task Instance\#\#\#

        The input image is the given image, the question is \{input question\}, the response is \{model response\}.\par\vspace{\baselineskip}

        \#\#\#Your Evaluation\#\#\#

        \end{tcolorbox}
\caption{Reasoning Evaluation Guideline.}
\label{fig:Reasoning Evaluation Guideline}
\end{figure*}

\begin{figure*}[htp]
        \centering
        \begin{tcolorbox}[title=Instruction Following Evaluation Guideline]
        Please act as an impartial and helpful evaluator for multimodal generation.
        Evaluate the quality of Instruction Following strictly based on the given evaluation criterion.\par\vspace{\baselineskip}
        Begin the evaluation by providing your analysis concisely and accurately, start with "Analysis:", and then on the next line, start with "Rating:" followed by your rating on a Likert scale from 1 to 5 (higher means better).
        You MUST keep to the strict boundaries of the evaluation criterion and focus solely on the issues and errors involved; otherwise, you will be penalized.
        Make sure you read and understand these instructions, as well as the following evaluation criterion carefully.\par\vspace{\baselineskip}

        \#\#\#Task Description\#\#\#
        
        Instruction Following: Given an image and text instructions, generate responses following the instructions.\par\vspace{\baselineskip}

        \#\#\#Evaluation Criterion\#\#\#

        Please evaluate mainly based on whether the response is informative, and whether the response contains any hallucination. Hallucination, in this context, refers to a situation where the MLLM generates a response that includes information not present or implied in the image or previous conversation. A hallucination could be a false claim about an object, action, emotion, or any other detail that is not grounded in the image.\par\vspace{\baselineskip}

        \#\#\#Task Instance\#\#\#

        The input image is the given image, the input instruction is \{input instruction\}, the response is \{model response\}.\par\vspace{\baselineskip}

        \#\#\#Your Evaluation\#\#\#

        \end{tcolorbox}
\caption{Instruction Following Evaluation Guideline.}
\label{fig:Instruction Following Evaluation Guideline}
\end{figure*}

\begin{figure*}[htp]
        \centering
        \begin{tcolorbox}[title=Human Evaluation Guideline of Rationale]
        Please evaluate the evaluation rationales from the following three perspectives.

        (1) Factual Accuracy

        Definition:
        Measures whether the rationale is factually correct with respect to the input (image and/or text) and free from hallucinations.
        
        Annotation Instructions:
        Annotators verify that all claims in the rationale are supported by the input and do not contradict observable content.
        
        Scoring:
        
        5 – Fully accurate; no factual errors.
        
        4 – Minor inaccuracies but overall correct.
        
        3 – Some noticeable unsupported claims.
        
        2 – Multiple factual errors.
        
        1 – Largely incorrect or hallucinated.

        (2) Relevance to Evaluation Criteria

        Definition:
        Measures whether the rationale properly addresses the specified evaluation criteria and justifies the assigned score.
        
        Annotation Instructions:
        Annotators assess whether the explanation focuses on the required evaluation dimensions and avoids irrelevant commentary.
        
        Scoring:
        
        5 – Directly and clearly aligned with the evaluation criteria.
        
        4 – Mostly aligned with minor irrelevance.
        
        3 – Partially aligned but incomplete or unfocused.
        
        2 – Weak connection to the evaluation criteria.
        
        1 – Not aligned with the evaluation criteria.

        (3) Coherence

        Definition:
        Measures the clarity, logical flow, and internal consistency of the rationale.
        
        Annotation Instructions:
        Annotators evaluate whether the explanation is well-structured, logically connected, and easy to follow.
        
        Scoring:
        
        5 – Clear, well-organized, and logically consistent.
        
        4 – Generally clear with minor structural issues.
        
        3 – Understandable but somewhat disorganized.
        
        2 – Difficult to follow.
        
        1 – Incoherent or contradictory.
        
        \end{tcolorbox}
\caption{Human Evaluation Guideline of Rationale.}
\label{fig:Human Evaluation Guideline of Rationale}
\end{figure*}

\end{document}